\documentclass{article}
\usepackage{spconf,amsmath,graphicx}

\usepackage{color}
\usepackage{xspace}
\usepackage{relsize}
\usepackage{caption}
\usepackage{subcaption}
\usepackage{multirow}
\usepackage{adjustbox}
\usepackage{float}
\usepackage{dblfloatfix}
\usepackage{tikz,pgfplots}
\pgfplotsset{compat=newest}
\usetikzlibrary{plotmarks, shapes, arrows, positioning, shadows, trees}
\usepackage{wrapfig}
\usepackage{booktabs}
\usepackage{setspace}

\newcommand\BLEU{\textsc{Bleu}\xspace}
\newcommand\TER{\textsc{Ter}\xspace}
\newcommand\WER{\textsc{Wer}\xspace}

\newlength\figureheight 
	\newlength\figurewidth

\title{Tight Integrated End-to-End Training for Cascaded Speech Translation}
\name{Parnia Bahar$^{1,2}$, Tobias Bieschke$^1$, Ralf Schl{\"u}ter$^{1,2}$ and Hermann Ney$^{1,2}$}
\address{
$^1$Human Language Technology and Pattern Recognition Group, Computer Science Department \\
RWTH Aachen University, 52074 Aachen, Germany, $^2$AppTek GmbH, 52062 Aachen, Germany \\
\texttt{\small \{bahar, schlueter, ney\}@cs.rwth-aachen.de, tobias.bieschke@rwth-aachen.de}
}

\begin{document}
%
\maketitle
\begin{abstract}
A cascaded speech translation model relies on discrete and non-differentiable transcription, which provides a supervision signal from the source side and helps the transformation between source speech and target text. Such modeling suffers from error propagation between ASR and MT models. Direct speech translation is an alternative method to avoid error propagation; however, its performance is often behind the cascade system. To use an intermediate representation and preserve the end-to-end trainability, previous studies have proposed using two-stage models by passing the hidden vectors of the recognizer into the decoder of the MT model and ignoring the MT encoder. This work explores the feasibility of collapsing the entire cascade components into a single end-to-end trainable model by optimizing all parameters of ASR and MT models jointly without ignoring any learned parameters. It is a tightly integrated method that passes renormalized source word posterior distributions as a soft decision instead of one-hot vectors and enables backpropagation. Therefore, it provides both transcriptions and translations and achieves strong consistency between them.
Our experiments on four tasks with different data scenarios show that the model outperforms cascade models up to 1.8\% in \BLEU and 2.0\% in \TER and is superior compared to direct models.

\end{abstract}

\section{Introduction \& Related Works} \label{sec:intro}

Conventional speech-to-text translation (ST) systems employ a two-step cascaded pipeline. The first component is an automatic speech recognition (ASR) system trained on speech-to-source data 
by which a spoken language utterance is first transcribed. The transcribed word sequence is translated by the second component, a machine translation (MT) system trained on source-to-target corpora.
The nearly-zero recognition error rate can hardly be expected in practice, and ASR errors are propagated to the MT model. 
The idea of tighter integrating these two models is not new \cite{casacubertaNOVVBGLMM04,matusov2005,casacubertaFNV08}.
It is arguable that the translation process might avoid some of these errors if multiple recognition hypotheses such as $N$-best list \cite{matusov05:pto}, lattices \cite{matusov05:pto,dyer2008:acl2008:lattice_translation,sperber2017neural} and confusion networks \cite{bertoldi2007:icassp2007:cn_st_dp,apptek_2018_st} are provided. 
In all such systems, explicit discrete intermediate transcriptions are generated, thus cascading requires doubled decoding time.

In contrast, recent direct models translate foreign speech without any need for transcriptions \cite{berard_2016_proof, goldwater_2017_noasr}.
Unlike conventional systems, each and every component of the direct model is trained jointly to maximize the translation performance, and it eliminates the two-pass decoding property.
Training such end-to-end models requires a moderate amount of paired translated speech-to-text data, which is not easy to acquire.
Previous works have mostly proposed remedies to the data scarcity problem of direct modeling such as leveraging weakly supervised data \cite{jia_2019_synthatic}, data augmentation \cite{bahar2019:iwslt-st-spectaug, McCarthy_2020}, multi-task learning \cite{weiss_2017_directly, st_kd_interspeech2019}, two-stage models \cite{ anastasopoulos_2018_tied_multitask, Sperber_19_attention_passing, Sung_2019} depending on intermediate transcripts, but optimized in an end-to-end fashion, pretraining different components of the model \cite{bansal_2019_pretraining_asr, Stoian_2020, bahar2019:st-comparison, bridging_the_gap} and multilingual setups \cite{inaguma2019multilingual, gangi_asru_multi}.

Assuming a realistic setup and not ignoring other available speech-to-source and source-to-target corpora, where cascade and direct models are trained on non-equal amounts of data, the performance of direct models is often behind cascaded systems.
The end-to-end methods either conduct translation without transcribing or suffer from inconsistency between transcriptions and translations \cite{sperber2020consistent}.
Transcriptions are essential in many applications and required to be displayed
together with translations to users.

Given that, two lines of research inspire our work. To 
\begin{itemize}
    \item have a realistic data condition and employ all types of training data, i.e. ASR, MT and direct ST corpora.
    \item provide both transcriptions and translations jointly with a strong consistency between them.
\end{itemize}

To this end, we propose a trainable cascade modeling, where an intermediate transcripts representation is used, but optimized in part through end-to-end training. 
Transcriptions create an inductive bias for better transformation between source speech and target text.
Since discrete representation is problematic for end-to-end training via backpropagation, \cite{ anastasopoulos_2018_tied_multitask, Sung_2019,Sperber_19_attention_passing} propose to pass hidden representations like attention vectors or decoder states from the ASR decoder to the MT encoder. 
The hidden neural states do not represent the word that has to be generated, rather a mix or a weighted representation of all words in the vocabulary. It is often a fuzzy representation to determine the correct word. 
Therefore, in this work, we pass the higher-level representation, i.e. renormalized source word posteriors. The approach is differentiable, can be optimized in part through end-to-end training, and offers a way to ease error propagation by conveying more information, i.e., uncertainty and potential other hypotheses. 

\begin{figure}
\centering
     \includegraphics[width=13.0cm,height=4.0cm,keepaspectratio]{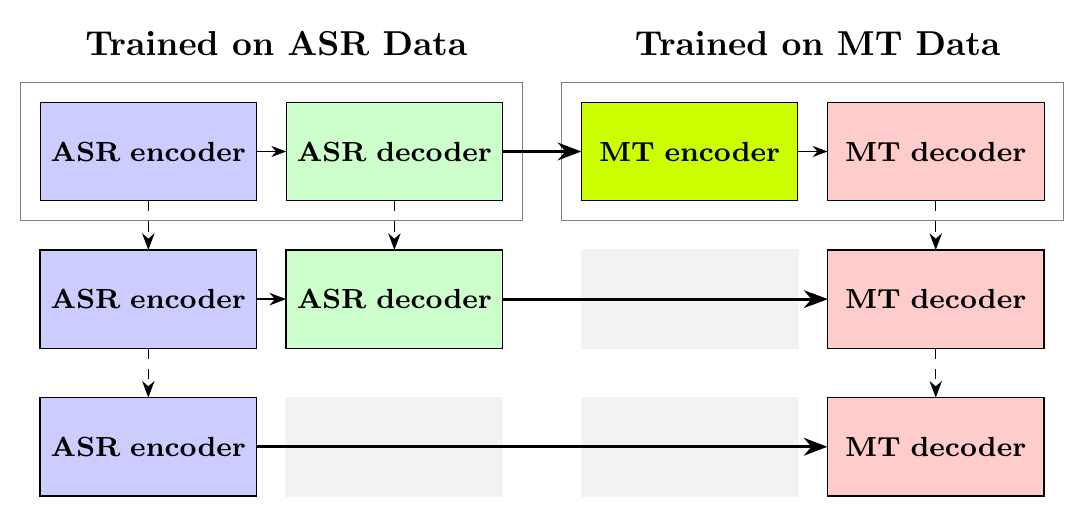}
  \caption{%
  From top to bottom, the figures depict a cascade system of ASR and MT models, an end-to-end trainable two-stage model where the MT encoder is ignored, an end-to-end direct training, where the ASR decoder and MT encoder are ignored. Our proposed integrated training model is framed to the first row, where we keep all parameters of the cascade system.}
\label{fig:pretraining}
\end{figure}

In addition, we take benefit from all components of pretraining. As illustrated in Figure \ref{fig:pretraining}, in the two-stage modeling \cite{anastasopoulos_2018_tied_multitask, Sung_2019, Sperber_19_attention_passing}, there is a second decoder, target text decoder on top of the ASR decoder. Such two-stage models discard many of the additional learned parameters, like the MT encoder (see the middle figure). 
The issue is more severe in the direct modeling (see the bottom figure) where
both the ASR decoder and MT encoder are completely neglected \cite{berard_2018_librispeech}. 
In the case of direct modeling, we do not consider almost half of the previously learned parameters. 
In contrast to those pretraining strategies, our model uses all
parameters of sub-models in the final end-to-end model.
There, it guarantees to be at least on par with the cascade system.

While many recent works investigated direct approaches to speech translation, this paper addresses the cascaded method and the tight integration of ASR and MT modules. Before the era of neural models, such a tight integration has been done using word lattices or word confusion networks \cite{DBLP:journals/csl/ManguBS00, matusov05:pto, Bertoldi_2005} at the interface between ASR and MT. This work revisits a similar idea of confusion networks by passing renormalized source word posteriors from ASR to the MT module.
The tightly integrated training allows exploiting speech-to-target training data, that might be available in some situations but cannot be used for cascaded model training.
Experimental results on different tasks show that it is superior despite its simplicity.

\section{Tight Integrated Training}

An intuitive notion in speech translation is that higher-level intermediate representations such
as transcriptions are beneficial for the final translation task since they provide a supervision signal from the source.
A problem with higher-level discrete representation is the non-differential property.  
The tight integrated model presents a novel architecture to entirely collapse ASR and MT models into a single network and jointly retrain all parameters using speech-to-target data.
The method is not a direct model as it depends on the intermediate representations. It is closer to the cascaded approach while maintaining end-to-end trainability.
The main idea behind our model is to adjust the posterior distribution of the ASR model (the first decoder) to be more peaked. A peaked distribution is closer to the one-hot vector representations which the encoder of the MT model expects. 
This contrasts with passing information at the lower-level of hidden representations from the ASR decoder into the MT encoder. 
The method is effective firstly because it is simple to apply as there is no need to change the architecture or training objective. Secondly, the ST task's usual data condition is a reasonable amount of speech-to-source parallel data, a lot of source-to-target bilingual data, and a little (low-resource) speech-to-target parallel data. 
Here are the core steps of our technique:

\begin{enumerate}
    \item Pretrain an ASR model with speech-transcribed data and an MT model with bilingual translation data
    
    \item Concatenate the ASR and MT models as used in the cascade setup
    
    \item To bridge the gap between traditional and integrated cascade models, we simulate a degree of ambiguities by passing the posterior distributions as a soft decision instead of a one-hot vector that is hard. As the MT model has been trained on one-hot representations, we sharpen the source word distribution by an exponent, $\gamma$, at each time step and renormalize the probabilities. The renormalized conditional probability of a source transcribed sequence $f_1^{J}$ with unknown length $J$ given an input observation (audio feature vectors) of variable length $T$, $x_1^{T}$ defines as
\begin{align}
p(f_j | f_1^{j-1} , x_1^T) = \frac{p^{\gamma}(f_j | f_1^{j-1} , x_1^T)}{\sum_{f'\in |V_F|}p^{\gamma}(f'_j | f_1^{j-1} , x_1^T)}
\label{eq:gamma}
\end{align}
where $|V_F|$ is the source vocabulary.
$\gamma=0$ leads to a uniform distribution over vocabulary. 
If we set $\gamma=1$, we get the posterior distribution and for $\gamma > 1$ a sharp distribution can be achieved such that $\gamma \to \inf$ represents almost one-hot representation. 
Another similar method is Gumbel softmax \cite{jangGP17_gumble} that we leave as our future work.
\item Continue training with the speech-to-target parallel corpus. If we skip this step and perform translation directly, it corresponds to a cascade model. 
We apply beam search on the ASR output to generate the best hypothesis during decoding and pass it to the MT. 
The model can be framed as in the first line of Figure \ref{fig:pretraining}.

\end{enumerate}

\section{Experiments}\label{sec:expriments}
\subsection{Datasets and Metrics} 
\label{sec:dataset}

We have performed our experiments on four speech translation tasks to explore different data conditions; IWSLT 2018 TED talks En$\to$De \cite{Cho_2014_ted}, MuST-C En$\to$De and En$\to$Es \cite{mustc} and LibriSpeech Audiobooks En$\to$Fr \cite{berard_2018_librispeech}.
The training data statistics are listed in Table \ref{tab:statistics}.
For the IWSLT TED En$\to$De and MuST-C tasks, we leverage additional speech-to-source and source-to-target data to build the cascaded system. For the MuST-C tasks, we increase the ASR data by almost eight times and do more careful filtering to select our MT data. For the MuST-C En$\to$De and En$\to$Es, the amount of ASR data is much more compared to the direct ST data.
Our data condition is unrealistic for LibriSpeech En$\to$Fr where we assume we have no additional indomain ASR or MT data available.

\noindent \textbf{IWSLT TED En$\to$De}:
We mainly follow the same data condition and data pipeline as explained in \cite{bahar2019:iwslt-st-spectaug,apptek_2018_st}.
We use the TED-LIUM corpus and the IWSLT speech translation TED corpus resulting in a total of 390h of speech data. 
Similar to \cite{apptek_2018_st}, we automatically recompute the provided audio-to-source-sentence alignments to reduce the problem of speech segments without a translation.
We randomly select a part of our segments as our cross-validation set and choose dev2010 and test2015 as our development and test sets with 888 and 1080 segments, respectively. 
As our bilingual text data, we use the TED, OpenSubtitles2018, Europarl, ParaCrawl, CommonCrawl, News Commentary, and Rapid corpora resulting in 32M sentence pairs after filtering noisy samples.
We apply byte pair encoding (BPE) \cite{sennrich_16_bpe} with $20$k operations jointly on both sides.

\noindent \textbf{MuST-C En$\to$De} and \textbf{MuST-C En$\to$Es:}
The setup is mainly based on \cite{bahar2020:iwslt:evaluation}.
The ASR model has been trained on a total of approx.~2300 hours of transcribed speech including EuroParl, How2, MuST-C, TED-LIUM, LibriSpeech 1000h, Mozilla Common Voice, and IWSLT TED corpora. We take this model as a general English ASR model and fine-tune on MuST-C En$\to$De and En$\to$Es data to obtain the final ASR models for these tasks, respectively. 
For MT training on En$\to$De, we utilize the parallel data allowed for the IWSLT 2020. After filtering the noisy corpora, namely ParaCrawl, CommonCrawl, Rapid and OpenSubtitles2018, we end up with almost 27M bilingual text sentences.
BPE symbols of size 5k and 32k are used on the English and German data.
Similarly, we take available data for En$\to$Es such as OpenSubtitles2018, News-Commentary, UN, TED, etc., filter the noisy data and yield around 48M sentence pairs. BPE size of 5k on the English part, respectively 32k symbols are used on the Spanish side.
The speech translation data is the MuST-C corpus as written in Table \ref{tab:statistics}.
We select MuST-C dev set, tst-HE, and tst-COMMON as our development and test sets, respectively. 

\noindent \textbf{LibriSpeech En$\to$Fr:}
As shown in Table \ref{tab:statistics}, we employ 100h of clean speech corresponding to 47.2k segments for ASR, MT and ST training.
The dev and test sets contain 2h and 4h of speech, 1071 and 2048 segments, respectively.
We apply 5k joint BPE units on both English and French texts.

For all tasks, 80-dimensional Mel-frequency cepstral coefficients (MFCC) features are extracted as our audio features.
For all tasks, we remove the punctuation from the transcriptions (i.e. the English text) and apply ASR-like noise \cite{bahar2020:iwslt:evaluation} and keep the punctuation on the target side. 
The same source BPE units are used on the ASR data, hence
the source vocabulary is shared between ASR and MT models. 
\texttt{Moses} toolkit \cite{koehn_07_moses}
is used for tokenization and true-casing. The evaluation is case-sensitive with punctuation for the IWSLT TED and MuST-C tasks while we lowercase for the LibriSpeech data.
The translation models are evaluated using \BLEU~\cite{papineni_02_bleu} computed by the official scripts of the WMT campaign,
i.e.
\texttt{mteval-v13a}\footnote{ftp://jaguar.ncsl.nist.gov/mt/resources/mteval-v13a.pl}. The results are identical to case-sensitive results computed by SacreBleu \cite{post-2018-call} except for LibriSpeech that is computed by Moses \texttt{multi-blue.pl} script on tokenized reference to be comparable with other works.
Normalized \TER~\cite{snover_06_ter} is computed by \texttt{tercom}\footnote{http://www.cs.umd.edu/~snover/tercom/}.
\WER~is computed by \texttt{sclite}\footnote{http://www1.icsi.berkeley.edu/Speech/docs/sctk-1.2/sclite.htm}.

\begin{table}[t]
\caption{Data statistics (number of sequences) and ASR results. $^1$: \WER on tst-HE and tst-COMMON respectively.}
\begin{center}
\scalebox{0.85}{%
\begin{tabular}{lrrr|rr}
\toprule
\multirow{2}{*}{\bfseries Task} & \multicolumn{3}{c}{\bfseries Data Condition}  & \multicolumn{2}{|c}{\bfseries \WER[$\downarrow$]}\\ \cline{2-4} \cline{5-6}
& \bfseries  ASR  &  \bfseries MT & \bfseries ST  & \bfseries  dev  &  \bfseries test\\ \midrule
IWSLT En$\to$De       &   263k  & 32M   & 171k  &  11.2  &  10.6 \\ 
MuST-C En$\to$De   &   1.9M  &  24M  & 222k & 11.8   & $^1$7.7/10.6\\ 
MuST-C En$\to$Es   &   1.9M  &  48M & 260k & 12.3 & $^1$7.3/10.4\\ 
LibriSpeech En$\to$Fr &   47.2k & 47.2k & 47.2k & 16.4  & 16.2\\
\bottomrule
\end{tabular}
}
\label{tab:statistics}
\end{center}
\end{table}

\subsection{Models} \label{sec:models}

All our models are based on long short-term memory (LSTM) \cite{hochreiter_97_LSTM} attention encoder-decoder \cite{bahdanau_15_attention} models.
We build an ASR model and an MT model concatenated to form the baseline cascaded system as well as a direct ST model.

\noindent \textbf{ASR Model:}
We map all subwords into embedding vectors of size 512.
The speech encoder is composed of 6 stacked bidirectional LSTM (biLSTM) layers equipped with 1024 hidden dimensions. We use two max-pooling operations between the first two biLSTM layers to reduce the audio sequence length by factors of 3 and 2 respectively, in total time reduction with a factor of 6. 
Layer-wise pretraining is used, where we start with two encoder layers, and iteratively add more layers until the 6th layer \cite{zeyer_18_returnn}. 
The subword-level decoder is a 1-layer unidirectional LSTM of size 1024 with a single head additive attention.
A variant of specAugment is also used \cite{bahar2019:iwslt-st-spectaug, park2019specaugment}. 

\noindent \textbf{MT Model:}
We also use a 6 layer biLSTM text encoder, with 1024 cells that used layer-wise pretraining.
Similar to the ASR decoder, we use a 1-layer LSTM with a cell size of 1024, with
single head attention. 512-dimensional embedding is used.

\noindent \textbf{ST Model:}
We use the same architecture to the ASR encoder and the same architecture to the MT decoder.

We train models using Adam update rule \cite{kingma_14_adam}, dropout of 0.3 \cite{srivastavad_14_dropout}, and label smoothing \cite{pereyra_2017_label_smoothing} with a ratio of 0.1.
We lower the learning rate with a decay factor in the range of $0.8$ to $0.9$ based on perplexity on the development set and wait for 6 consecutive checkpoints. The maximum sequence length is set to 75 tokens. The batch sizes are chosen to fit to the memory of the GPUs.
We use a beam search with a size of 12 everywhere.
For the tight integrated model, we vary $\gamma$ during decoding and once we obtain a proper value, we use it for all decoding jobs. Label smoothing is disabled and a learning rate of $0.00001$ to $0.00008$ is used.
We use our in-house implementation in
\texttt{RETURNN} \cite{zeyer_18_returnn}.
The code\footnote{https://github.com/rwth-i6/returnn} 
and the configurations of the setups are available online\footnote{https://github.com/rwth-i6/returnn-experiments/}.

\begin{table}[t]
\caption{\BLEU~$^{[\%]}$ and \TER~$^{[\%]}$ scores for various $\gamma$ values on cascade of ASR and MT models during decoding.}
\label{tab:gamma_results}
\begin{center}
\scalebox{1.0}{%
\begin{tabular}{lllll}
\toprule
\multirow{3}{*}{\bfseries $\gamma$} & \multicolumn{2}{c}{\bfseries IWSLT} & \multicolumn{2}{c}{\bfseries LibriSpeech}  \\ \cline{2-5}
 & \multicolumn{2}{c}{\bfseries dev2010} & \multicolumn{2}{c}{\bfseries dev}  \\ \cline{2-5}
 & \bfseries  \BLEU  & \bfseries  \TER &  \bfseries \BLEU  &  \bfseries \TER\\ 
\midrule

0.5 & - & - & - & - \\
0.9 & 12.9 & 79.7 & 22.6 & 66.4 \\
1.0 & 21.9 & 64.6 & 22.8 & 65.6\\
1.5 & 26.0 & 56.6 & 23.0 &  65.4\\
\textbf{2.0} & 26.0 & 56.7 & 23.0 &  65.4\\
4.0 & 26.0 & 56.7 & 23.1 &  65.3\\
32 & 25.9 & 56.8 & 23.0 & 65.5\\
128 & 25.9 & 56.8 & 23.1 &  65.5\\
1024 & 25.9 & 56.8 & 23.0 &  65.5\\

\bottomrule
\end{tabular}
}
\end{center}
\end{table}

\begin{table}[t]
\caption{Results on IWSLT En$\to$De. $^1$: based on our implementation and data.}
\label{tab:st_results_iwslt}
\begin{center}
\scalebox{0.80}{%
\begin{tabular}{lrllll}
\toprule
\multirow{2}{*}{\bfseries Model} & \multicolumn{2}{c}{\bfseries dev2010 } & \multicolumn{2}{c}{\bfseries test2015 }  \\ \cline{2-5}
& \bfseries  \BLEU  & \bfseries  \TER &  \bfseries \BLEU  &  \bfseries \TER\\ 
\midrule
\multicolumn{5}{l}{\bfseries other works}\\ 
IWSLT 2018 winner - direct \cite{liu_2018_iwslt} & & & 20.1 &  \\
IWSLT 2018 winner - cascade  \cite{liu_2018_iwslt} & & & 26.0 &  \\
\midrule
\multicolumn{5}{l}{\bfseries this work}\\ 

text MT & 29.8  &  50.7  &  31.7  & 51.7 \\

cascade  &  26.2 & 56.1 & 27.3 & 57.5 \\ 

direct end2end  + pretraining       & 22.7 & 59.6 & 21.6 & 64.2 \\ 

multi-task (w MT)\cite{weiss_2017_directly}$^1$ + pretraining   &21.3 & 62.0 & 20.6 & 65.9 \\

multi-task (w ASR)\cite{weiss_2017_directly}$^1$ + pretraining   & 21.1 & 61.8 & 19.2 & 67.9  \\

attention passing\cite{Sperber_19_attention_passing}$^1$  + pretraining  & 19.6 & 62.9 & 19.6 & 65.9 \\ 

tied multitask\cite{anastasopoulos_2018_tied_multitask}$^1$ + pretraining  & 20.4 & 61.7 & 20.1 & 65.8 \\

tight integrated cascade & \textbf{26.8} & \textbf{55.1} & \textbf{28.1} & \textbf{56.7} \\

\bottomrule
\end{tabular}
}
\end{center}
\end{table}

\section{Results}

Table \ref{tab:statistics} presents the performance of the ASR models.
On the test sets, we achieve 10.6\% on IWSLT TED tst2015, 7.7\% and 10.6\% on MuST-C tst-HE and tst-COMMON on En$\to$De, 7.3\% and 10.4\% on En$\to$Es and 16.2\% \WER on LibriSpeech respectively.
Obtaining the best hypothesis for each utterance by the ASR model, we pass it into the MT model.
To this end, we concatenate the ASR and MT networks end-to-end, and as a sanity check, we pass the one-hot vectors as the outputs of the ASR into the MT model. 
This has led to the exact same result as the vanilla cascade approach.
This step needs to have a shared source vocabulary on the ASR and MT side, unlike the cascade system. 

\subsection{Effect of $\gamma$ in Passing Posteriors}

We then attempt to shed light on the question of which $\gamma$ values in Eq. \ref{eq:gamma} make the distribution to be more peaked.
Thus, during decoding, instead of passing the one-hot vectors, we pass renormalized posterior distributions. By varying the $\gamma$ value, we try to adjust the distribution to be closer to one-hot representations expected by the MT embedding layer. Table \ref{tab:gamma_results} lists different values versus translation scores on dev sets. 

As shown, for a small value of $\gamma=0.5$, we yield no reasonable translation performance, as the probability of words is uniformly distributed over vocabulary.
We found that $\gamma=2$ is already on par with the results of the cascade approach on the dev sets.
The performance is barely changed as the distribution is already sharp enough and a further increase of $\gamma$ exponent does not change the distribution.
The MT encoder is trained to see one-hot vectors and does not know how to deal with the probability distribution like posteriors. 
By applying sharpening, we focus more on the word with the highest probability, resulting in similar greedy search performance.
We note that we observe the same behavior on the \WER performance of the ASR model.
We also highlight that in all experiments, we use $\gamma=1$ in training and $\gamma=2$ in decoding.

\begin{table*}[tp]
\begin{center}
\caption{Results measured in \BLEU~$^{[\%]}$ and \TER~$^{[\%]}$ on the MuST-C task.}
\label{tab:st_results_mustc}
\scalebox{0.9}{%
\begin{tabular}{lllllllll}
\toprule
\multirow{3}{*}{\bfseries Model} & \multicolumn{4}{c}{\bfseries MuST-C En$\to$De} & \multicolumn{4}{c}{\bfseries MuST-C En$\to$Es} \\ \cline{2-9}
& \multicolumn{2}{c}{\bfseries tst-HE}   & \multicolumn{2}{c}{\bfseries tst-COMMON } & \multicolumn{2}{c}{\bfseries tst-HE}   & \multicolumn{2}{c}{\bfseries tst-COMMON } \\ \cline{2-5} \cline{6-9}
& \bfseries  \BLEU  & \bfseries  \TER &  \bfseries \BLEU  &  \bfseries \TER & \bfseries  \BLEU  & \bfseries  \TER &  \bfseries \BLEU  &  \bfseries \TER \\ 
\midrule
\multicolumn{8}{l}{\bfseries other works}\\ 

direct   \cite{Gangi_inter19}   &  & & 17.3 & & & & 20.8& \\
multilingual   \cite{gangi_asru_multi}   & & &  17.7  & & & & 20.9&\\
\midrule
\multicolumn{8}{l}{\bfseries this work}\\ 

text MT &  27.6 & 55.7 &  29.6 & 51.4 & 41.9 	& 	41.8 	&	34.1 	&	50.0\\
cascade  & 25.0 & 59.2 & 25.9 &  56.2 & 37.6 & 46.5 &	30.2 &	55.4\\ 
direct+pretraining    & 24.4 & 59.9 & 25.1& 56.9 & 35.2	& 49.1	& 28.7 &	56.1 \\ 
tight integrated cascade & \textbf{26.8} & \textbf{57.2} & \textbf{26.5}& \textbf{54.8} & \textbf{38.0} &	\textbf{46.4} &	\textbf{30.6} &	\textbf{55.0}\\
\bottomrule
\end{tabular}
}
\end{center}
\end{table*}

\begin{table}[t]
\begin{center}
\caption{Results measured in \BLEU~$^{[\%]}$ and \TER~$^{[\%]}$ on LibriSpeech En$\to$Fr. The evaluation is based on lowercase tokenized \BLEU computed by Moses \texttt{multi-blue.pl} script.
}
\label{tab:st_results_libri}
\scalebox{0.85}{%
\begin{tabular}{lrllll}
\toprule
\multirow{2}{*}{\bfseries Model} & \multicolumn{2}{c}{\bfseries dev } & \multicolumn{2}{c}{\bfseries test }  \\ \cline{2-5}
& \bfseries  \BLEU  & \bfseries  \TER &  \bfseries \BLEU  &  \bfseries \TER\\ 
\midrule
\multicolumn{5}{l}{\bfseries other works}\\ 
direct+pretraining \cite{berard_2018_librispeech}        & &  & 13.3 &\\
direct+knowledge distillation  \cite{st_kd_interspeech2019} & &  & 17.0 & \\
multilingual   \cite{InagumaDKW19}   & &  & 17.6 & \\
direct+curriculum pretrain  \cite{curculum_pre} & &  & 18.0 & \\
direct+synthetic data  \cite{McCarthy_2020} & &  & 22.4 & \\

\midrule
\multicolumn{5}{l}{\bfseries this work}\\ 

text MT & 25.2 & 62.6 & 22.9 & 64.4 \\
cascade  & 23.3 & 65.2 & 21.4 & 66.4 \\ 
direct+pretraining  & 20.6 & 66.9 & 21.3 & 66.0 \\ 
tight integrated cascade & \textbf{23.9} & \textbf{64.0} & \textbf{21.4} & \textbf{65.4} \\

\bottomrule
\end{tabular}
}
\end{center}
\end{table}

\subsection{Translation Performance}
Once we obtain $\gamma$, we tighten the ASR and MT models and train the coupled model using speech-to-target data in an end-to-end fine-tuning fashion.
Table \ref{tab:st_results_iwslt} presents the IWSLT task results and a comparison with other works.
As expected, due to the error propagation of the cascade model, its performance is significantly lower than that of the pure text MT model.
The ASR and MT models built the cascade baseline are used for pretraining the encoders and decoders of the direct model equipped with an adapter layer \cite{bahar2019:st-comparison} as well as constructing multi-task models.
As shown, the direct model's performance is far behind the cascade model when using other available ASR and MT data. 
In multi-task learning, auxiliary data helps only up to some degree. 
The optimization of multi-task models is extremely dependent on the ST data.
Our results also indicate that the tightly integrated model outperforms the direct model by 6.5\% in \BLEU and 7.5\% in \TER on test2015, respectively improves the results of the cascade model by 0.8\% in both \BLEU and \TER.
We also compare our method to the similar works in which two-stage modeling has been proposed by passing the attention context vectors \cite{Sperber_19_attention_passing} or the decoder hidden states \cite{anastasopoulos_2018_tied_multitask} of the recognizer into the decoder of the MT model and ignoring the MT encoder. 
As shown in Table \ref{tab:st_results_iwslt}, our tight integration modeling is superior in comparison to them by a large margin. We again note that in such modeling, (almost) one-fourth of all parameters are disregarded while fine-tuning (middle row of Figure \ref{fig:pretraining}). 

We also report the results on the MuST-C and LibriSpeech tasks in Tables \ref{tab:st_results_mustc} and \ref{tab:st_results_libri} respectively, along with the comparison to the literature. For the MuST-C tasks, we use 2300h of ASR data, which is almost eight times more than those used in the IWSLT task, and the amount of ST data is also larger compared to the IWSLT task.
With this data condition and pretraining, we close the gap between the cascade and direct models and get the difference of 0.7\% in \BLEU and \TER on average of two test sets on En$\to$De translation.
Again, the tight integrated model outperforms the cascade system by 1.8\% and 0.6\% in \BLEU and 2.0\% and 1.4\% in \TER on tst-HE and tst-COMMON, respectively. It also outperforms the direct model.
The same behavior can be seen on MuST-C En$\to$Es with a gain of 0.4\% in \BLEU on average.
On LibriSpeech, we only use the ST data for model training limited to 47.2k segments. As presented in Table \ref{tab:st_results_libri}, the direct model reaches the cascade model on the test set, even yields better \TER score.
There, the tight model is also comparable to the cascade system since the model parameters are trained using already seen samples, and fine-tuning does not bring improvements in terms of \BLEU, though it helps \TER by 1\%.
A larger improvement is achieved on the dev set.
Even on a low-resource data condition, the tight integrated model guarantees at least the performance of the cascaded system.

\begin{table*}[th]
\begin{center}
\caption{Impact of subtask training measured in \BLEU~$^{[\%]}$, \TER~$^{[\%]}$ and \WER~$^{[\%]}$.}
\label{tab:st_results_iwslt_subtask}
\scalebox{1.0}{%
\begin{tabular}{llll|lll|lll}
\toprule
\multirow{3}{*}{\bfseries Model} & \multicolumn{3}{c|}{\bfseries IWSLT En$\to$De} & \multicolumn{3}{c}{\bfseries MuST-C En$\to$Es } & \multicolumn{3}{|c}{\bfseries LibriSpeech En$\to$Fr } \\ \cline{2-10}
& \multicolumn{3}{c}{\bfseries test2015 } & \multicolumn{3}{|c}{\bfseries tst-COMMON } & \multicolumn{3}{|c}{\bfseries test}  \\ \cline{2-5} \cline{6-10}
& \bfseries  \BLEU  & \bfseries  \TER & \bfseries  \WER &  \bfseries \BLEU  &  \bfseries \TER & \bfseries  \WER & \bfseries  \BLEU  & \bfseries  \TER & \bfseries  \WER \\ 
\midrule

tight integrated cascade &  28.1 & 56.7 & 13.0 &  30.6 & 55.0 & 11.2 & 21.4 & 65.4 & 16.0\\
\quad frozen ASR (dec \& enc) &  28.4 & 56.3 & 10.6 & 30.9 & 54.8 & 10.4 & 21.2& 65.7& 16.2\\
\quad frozen ASR dec & 28.4 & 56.2 & 10.7& 30.7 & 54.7 & 10.6& 21.3& 65.7& 16.0\\
\quad frozen MT enc & 28.1 & 56.6 & 12.9 & 30.0 & 55.5 & 11.4& 21.3& 66.2& 16.0\\
\quad frozen ASR dec \& MT enc & 28.4 & 56.3 & 10.6 & 30.4& 54.9 & 11.4 & 21.5& 66.1& 16.0\\

\bottomrule
\end{tabular}
}
\end{center}
\end{table*}

\subsection{Effect of Subtask Training}
The tight integrated training can be considered as a multi-task setup of recognition and translation tasks jointly, where we aim to improve the generalization performance of the ST task. 
In this case, we need to compromise between multiple tasks, and that might lead to a sub-optimal solution for the entire optimization problem. 
Hence, we also explore different strategies by training sub-components of the network and freezing the rest of the parameters.
One natural question is how to guarantee the initial \WER or even improve it at the end? This is particularly important for those applications where we require to generate both transcriptions and translations. 
Both translation and recognition results are listed in Table \ref{tab:st_results_iwslt_subtask}. 
Training all parameters results in worse \WER for the IWSLT and MuST-C tasks (cf. Table \ref{tab:statistics}).
When we freeze all ASR parameters and only update the MT parameters during training, it favors the ASR task as we do not update the ASR parameters and guarantee exactly the same \WER. 
Due to better \WER compared to line 1, we also achieve better \BLEU and \TER scores.
This implies a high correlation between transcriptions and translations that indicates a strong consistency between them.
We believe this consistency is an inherent property of our tight model as the translations depend on the transcriptions.

This makes different results on LibriSpeech with limited training data. As shown, joint training of all components slightly helps both recognition and translation performances. However, the correlation between the results stays consistent.
Fixing the ASR decoder also does not hurt the recognition error, whereas freezing MT encoder performs worse in translation due to higher recognition scores on the high-resource tasks.
In order to simulate the behavior of the direct modeling, we fix both the ASR decoder and MT encoder and update the rest of the parameters led to mixed results.
Comparing sub-task training to the other two-stage models listed in Table \ref{tab:st_results_iwslt}, one can conclude that it is beneficial to use all pretrained parameters of the cascaded model in a joint training fashion, though freeze some.

\subsection{Effect on training data size}
As shown in the tables, the tight integration helps the \BLEU and \TER scores in almost all scenarios irrespective of the amount of training data. However, the degree of performance improvement seems to be greatly dependent on the corpus. 
The largest gain is where we leverage a better data condition on the ASR and MT part and a reasonable amount of direct ST data. 
The minimal gain on test set occurs for the LibriSpeech task where we utilize nor an additional ASR data neither MT. In this scenario, perhaps the model has been already reached its capacity to learn, and fine-tuning stage using tight integration has smaller effect.

We believe it is crucial to tighten the ASR and MT models as an end-to-end trainable network and optimize all parameters jointly. 
Understanding the weaknesses and strengths of such models 
allows us to determine future research directions.

\section{Conclusion} \label{sec:conclusion}
We have proposed a simple yet elegant tight model enabling a fully end-to-end training for the cascade of ASR and MT models. 
The model takes all types of training data and provides both transcriptions and translations jointly with consistency based correlation between them. However, a quantitative measurement of consistency has been left as our future work.
Instead of one-hot representations, we have passed renormalized posterior distributions into the MT model, where we backpropagate the error.
Besides the data available for ASR and MT training, the method leverages the speech-to-target data to fine-tune a model whose network's parameters have
been separately pretrained on ASR and MT tasks. 
We demonstrate that this approach outperforms both the cascade and direct ST models on speech translation tasks. 
Freezing the ASR model's parameters guarantees the same recognition accuracy.
To convey more information and pass more ambiguities of the recognition system, we plan to adapt confusion networks in the tight integrated training in the future. We also want to explore a non-autoregressive model on the ASR output to avoid the need for beam search during decoding.

\section{Acknowledgements}
\label{sec:acknowledgements}
\begin{wrapfigure}{l}{0.10\textwidth}
\vspace{-8mm}
    \begin{center}
      \includegraphics[width=0.12\textwidth]{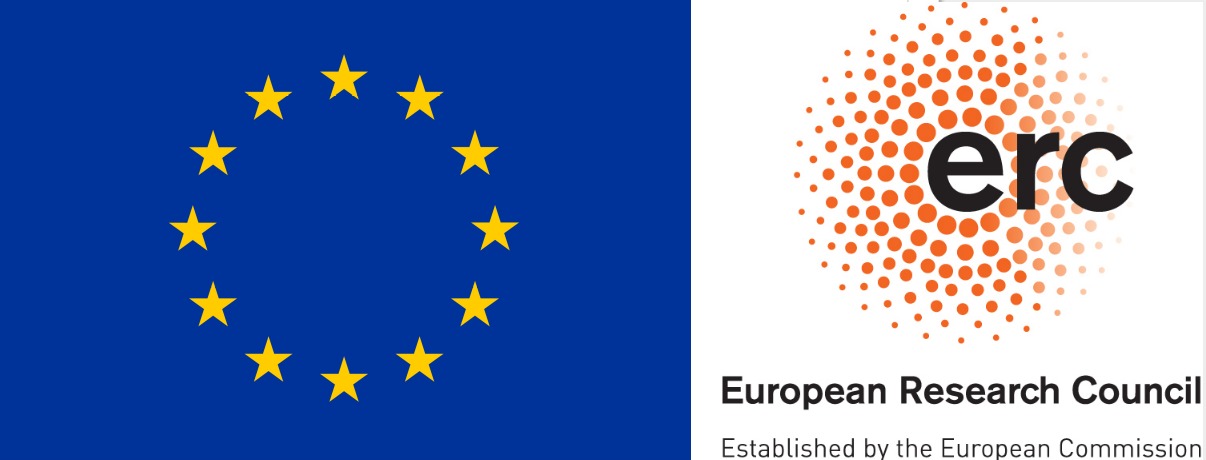} \\
      \vspace{2mm}
      \includegraphics[width=0.12\textwidth]{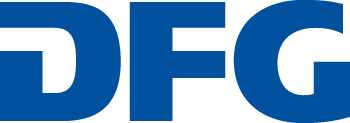}
    \end{center}
\vspace{-7mm}
\end{wrapfigure}
This work has received funding from the European Research Council (ERC) under the European Union's Horizon 2020 research and innovation programme (grant agreement No 694537, project "SEQCLAS"), the Deutsche Forschungsgemeinschaft (DFG; grant agreement NE 572/8-1, project "CoreTec") and from a Google Focused Award. The work reflects only the authors' views and none of the funding parties is responsible for any use that may be made of the information it contains.



\bibliographystyle{IEEEbib}



\let\OLDthebibliography\thebibliography
\renewcommand\thebibliography[1]{
  \OLDthebibliography{#1}
  \setlength{\parskip}{0pt}
  \setlength{\itemsep}{0pt plus 0.07ex}
}

\bibliography{refs}
\end{document}